# Fast Dual-Regularized Autoencoder for Sparse Biological Data

Aleksandar Poleksic
University of Northern Iowa, Cedar Falls, IA, USA.
`aleksandar.poleksic@uni.edu`

**Abstract**

Relationship inference from sparse data is an important task with applications ranging from product recommendation to drug discovery. A recently proposed linear model for sparse matrix completion has demonstrated surprising advantage in speed and accuracy over more sophisticated recommender systems algorithms. Here we extend the linear model to develop a shallow autoencoder for the dual neighborhood-regularized matrix completion problem. We demonstrate the speed and accuracy advantage of our approach over the existing state-of-the-art in predicting drug-target interactions and drug-disease associations.

## 1 Introduction

Algorithms for sparse matrix completion are used in recommender systems to predict user preferences to items such as news, movies, or songs [1]. The same methods can be successfully applied in other fields, for instance in systems biology to predict gene-disease associations or in computational systems pharmacology to predict adverse drug reactions [2] and to repurpose FDA approved drugs [3]. Matrix completion is the task of filling out missing entries in an observed sparse matrix. A low rank solution to matrix completion problem can be obtained via matrix factorization, a technique that approximates the input sparse matrix as a product of two lower dimensional matrices of users' and items' latent vectors [4]. Despite efforts to develop more sophisticated techniques, such as the methods based on artificial neural networks [5], matrix factorization remains the method of choice in recommender systems due to its efficiency and high accuracy [6]. Surprisingly however, a simple linear model, called EASE [7], has been recently shown to outperform other widely used recommender systems algorithms [7,8]. Aside from having competitive ranking accuracy, EASE is extremely fast as it based on an explicit, closed form solution. The method can be viewed as a shallow autoencoder that employs a single hyperparameter to optimize item-item weights, which are then applied to fill out the missing entries of the input sparse matrix. Recently, Jeunen *et al.* extended the EASE algorithm to incorporate side-information encoded in the tag-item matrix [9,10].

Here we show how the approaches in [7,9,10] can be adopted and further generalized to improve the ranking accuracy of recommender systems algorithms and particularly algorithms for biological relationship inference. To predict associations of elements from two biological domains $\mathcal{L}_1$ (which can be thought of as the user domain) and $\mathcal{L}_2$ (thought of as the item domain), we utilize homophily information in the form of pairwise similarities between elements of $\mathcal{L}_1$ and pairwise similarities between elements of $\mathcal{L}_2$. This technique is further extended to incorporate multiple other sources of information encoded in a heterogenous biological network. We show that our dual-regularized autoencoder, called DUET, yields more accurate classification scores when compared to both the EASE method and a state-of-the-art logistic matrix factorization technique.

This paper is organized as follows. In section 2.1, we present a straightforward extension of the EASE method that computes and utilizes user-user weights. In section 2.2, we discuss a special form of the procedure given in [9,10], namely one capable of incorporating homophily information into the learning process. Section 2.3 presents a further generalization of methods in [7,9,10] that incorporates multiple sources of side information to better model relationships between elements of two biological domains. Finally, in the Results section, we demonstrate the benefits of our approach in the example settings of predicting drug-target and drug-disease associations.

## 2 Methods

### 2.1 Computing user-user weights

To fill out the missing entries of a binary, sparse user-item interaction matrix $X$, the EASE algorithm learns the item-item weight matrix $B$ by solving

$$\min_B \|X - XB\|_F^2 + \lambda_1 \|B\|_F^2 \qquad (1)$$
$$\text{s.t. } diag(B) = 0,$$

where $\| \ \|_F^2$ denotes the Frobenius norm, $diag(B)$ is the vector of diagonal elements of $B$ and $\lambda_1$ is a trainable parameter. As noted in [7], the constraint $diag(B) = 0$ allows the model to generalize, giving rise to the completed user-item interaction matrix $XB$.

We note that the above approach does not take advantage of user-user weighs, although they can be easily learned by solving

$$\min_U \|X - UX\|_F^2 + \lambda_2 \|U\|_F^2 \qquad (2)$$
$$s.t. \ diag(U) = 0.$$

Specifically, to solve (2) one can apply the procedure in [7] to the transpose of $X$. With both weight matrices $U$ and $B$ in place, the completed matrix of user-item interactions can be defined as the average of $UX$ and $XB$.

### 2.2 Utilizing homophily information

A more accurate model can be designed in cases of available user-user and item-item similarity scores. For instance, in the setting of drug-target interaction prediction (drugs and proteins are here viewed as users and items, respectively), drug-drug similarities may be computed as Tanimoto scores (Jaccard indices) between drug feature vectors [11]. Protein-protein similarity scores may be defined as the pairwise similarity scores between the protein amino-acid sequences, which can be computed using the Smith-Waterman [12] or similar algorithms.

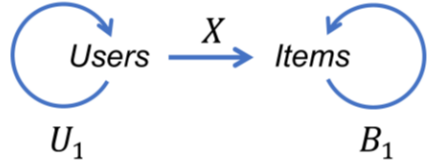

**Fig. 1.** Dual-regularized EASE algorithm (DUET) takes advantage of user-user ($U_1$) and item-item ($B_1$) similarity scores (where available) to better predict user-item associations.

We recall that the original EASE method learns item-item weights $B$ solely based upon the user-item interactions. In contrast, our DUET method is capable of learning both, user-user and item-item weights ($U$ and $B$, respectively) based upon not only the user-item interactions, but also on the pairwise similarities of users, and the pairwise similarities of items (labelled $U_1$ and $B_1$ in Fig. 1, respectively). The actual algorithm is presented as part of a more general procedure in the next section.

## 2.3  Utilizing multiple sources of side information

While the underlying idea behind the DUET algorithm has been described and used in a different setting in [9,10], here we describe a generalization of the method in [9,10], namely the one that takes account of other types of associations between users, items, and other relevant entities, as shown in Fig. 2. We note that the network in Fig. 1 is a special case of the network in Fig. 2, where $m = n = 1$, $J_1 = Users$ ($U_1$ = pairwise similarities of users), and $L_1 = Items$ ($B_1$ = pairwise similarity of items).

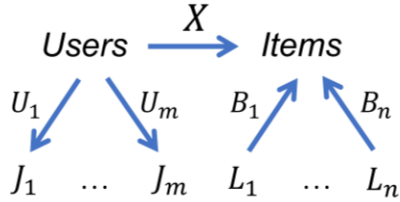

**Fig. 2.** Generalized DUET algorithm takes account of relationships between users and entities from multiple domains $J_1, \dots, J_m$. It also takes account of relationships between items and entities from multiple related domains $L_1, \dots, L_n$.

Specifically, given the heterogenous network in Fig. 2, we first optimize item-item weights $B$ by solving

$$\min_{B} \|X - XB\|_F^2 + \sum_i \beta_i \|B_i - B_i B\|_F^2 + \lambda_1 \|B\|_F^2 \tag{3}$$
$$s.t.\ diag(B) = 0,$$

where $\{B_i\}_1^n$ are the input matrices representing associations between items and elements from multiple other domains $\{L_i\}_{i=1}^n$. For example, in case where items represent movies, an example domain may consist of a set of movie festivals. A movie is "associated" with a festival if it won a festival's award. Other biological examples are provided later in this article.

We optimize user-user weights $U$ by solving

$$\min_U \|X - UX\|_F^2 + \sum_i \gamma_i \|U_i - UU_i\|_F^2 + \lambda_2 \|U\|_F^2 \qquad (4)$$
$$s.t.\ diag(U) = 0.$$

where $\{U_i\}_1^m$ are the sparse matrices representing associations of users and elements from multiple other domains $\{J_i\}_{i=1}^m$.

As in [7], a closed form solution of (3) can be obtained by minimizing the error function

$$E = \|X - XB\|_F^2 + \sum_i \beta_i \|B_i - B_i B\|_F^2 + \lambda_1 \|B\|_F^2 + 2\gamma^T diag(B), \qquad (5)$$

where $\gamma = (\gamma_1, \dots, \gamma_n)^T$ are Lagrange multipliers and $\{\beta_i\}$ and $\lambda_1$ are trainable parameters. Setting $\partial E/\partial B = 0$ we have

$$(X^T X + \sum_i \beta_i B_i^T B_i + \lambda_1 I)B = X^T X + \sum_i \beta_i B_i^T B_i - diagMat(\gamma), \qquad (6)$$

where $diagMat(\gamma)$ is the diagonal matrix with the elements of $\gamma$ on the main diagonal. Hence,

$$B = P(P^{-1} - \lambda_1 I - diagMat(\gamma)), \qquad (7)$$

where

$$P = (X^T X + \sum_i \beta_i B_i^T B_i + \lambda_1 I)^{-1}. \qquad (8)$$

Following further the general approach in [7],

$$B = I - P(\lambda_1 I + diagMat(\gamma)), \qquad (9)$$

or equivalently

$$B = I - P \cdot diagMat(\bar\gamma), \qquad (10)$$

where

$$\bar\gamma = \lambda_1 \vec{1} + \gamma, \qquad (11)$$

and where $\vec{1}$ is the vector of 1s. Since $diag(B) = 0$, it follows that

$$\vec{1} - diag(P) \odot \bar\gamma = 0, \qquad (12)$$

and thus

$$\bar\gamma = \vec{1} \oslash diag(P), \qquad (13)$$

where $\odot$ and $\oslash$ denote elementwise product and division, respectively. Substituting (13) into (10) we get

$$B = I - P \cdot diagMat\left(\vec{1} \oslash diag(P)\right). \tag{14}$$

In a similar way, we can compute user-user weights $U$ in (4) as

$$U = I - diagMat\left(\vec{1} \oslash diag(Q)\right) \cdot Q, \tag{15}$$

where

$$Q = (XX^T + \sum_i \gamma_i U_i U_i^T + \lambda_2 I)^{-1}, \tag{16}$$

and where $\{\gamma_i\}$ and $\lambda_2$ are trainable parameters.

## 3 Results

### 3.1 DrugBank benchmark

We first assessed the added value of our method on the task of predicting drug-target interactions. Our experimental setup uses the DrugBank database consisting of 9,881 interactions between 1482 FDA-approved drugs and 1408 target proteins [13]. Sparse interaction matrix $X$ and the matrices of protein-protein ($B_1$) and drug-drug ($U_1$) similarity scores used in our benchmarking experiment are available at *www.cs.uni.edu/~poleksic/drugbank_files.tar.gz*. This experimental setup corresponds to the special case of biological network in Fig. 2, namely the one depicted in Fig. 1.

To provide insight into the speed and accuracy of our algorithm, we compare it with the EASE method and the in-house logistic matrix factorization (MF) algorithm COSINE [4]. The COSINE method generalizes some popular matrix factorization algorithms [14,15] by utilizing weights on drug-target interactions. However, since drug-target interaction weights are not available in our test setting, the version of the COSINE algorithm benchmarked here is equivalent to the well-known NRLMF method described in [15]. In numerous studies published over the last decade, the logistic matrix factorization algorithms are shown to compare favorably to other state-of-the art methods for predicting drug-target interactions.

**Table 1.** DrugBank benchmark

|  | AUPR | NDCG100 | PREC50 | PREC100 |
|---|---|---|---|---|
| **MF0** | 0.483±0.002 | 0.988±0.004 | 0.991±0.005 | 0.985±0.004 |
| **EASEd** | 0.451±0.002 | 0.989±0.006 | 0.995±0.006 | 0.987±0.006 |
| **EASEt** | 0.487±0.003 | 0.987±0.015 | 0.976±0.008 | 0.967±0.003 |
| **EASEdt** | *0.492±0.003* | *0.992±0.002* | *0.997±0.002* | *0.991±0.002* |
| **MF** | 0.549±0.004 | 0.982±0.002 | 0.992±0.004 | 0.979±0.003 |
| **DUET** | **0.580±0.002** | **0.997±0.001** | **1.000±0.000** | **0.997±0.001** |

Our benchmarking procedure uses three rounds of the classical 5-fold cross-validation (CV). In each CV round, the input drug-target association matrix $X$ is randomly split into 5 groups. Each group is used once as test data, while the remaining four groups represent training data. The final classification scores (AUPR, $NDCG_{100}$, PREC@50, and PREC@100) are computed by averaging the classification scores obtained across different CV rounds.

Table 1 shows the performance of six different methods in the DrugBank benchmark after a thorough optimization of the methods' parameters. The first four methods, namely MF0, EASEd, EASEt, and EASEdt do not utilize any drug-drug or target-target similarity information. Specifically, MF0 is the barebone logistic matrix factorization algorithm (as implemented in COSINE) that has no access to homophily information. EASEt denotes the original EASE method which learns and applies target-target weights to predict missing drug-target interaction probabilities while EASEd optimizes drug-drug weights (section 2.1). The EASEdt method combines the results of EASEt and EASEd by averaging their prediction scores, exactly as described in section 2.1. The MF method is the full-blown matrix factorization algorithm that utilizes homophily information [4,15]. Finally, as previously noted, the DUET algorithm is simply EASEdt but guided by the input drug-drug and target-target similarity scores (as described in section 2.2 and shown in Fig. 1).

As seen in Table 1, EASEdt significantly outperforms (t-test p-value less than 0.01) not only MF0 but also both EASEt and EASEd, underlying the importance of computing and utilizing both drug-drug and target-target weights for more accurate relationship inference. More importantly, the DUET method significantly outperforms all other algorithms across all classification metrics. The advantage of DUET over EASEdt emphasizes the importance of utilizing side information from both drugs and targets in predicting drug-target associations.

## 3.2 Drug repurposing benchmark

In our second benchmark, we assessed the accuracy of the DUET algorithm in predicting drug-disease associations from a small drug-gene-disease network. This network is shown in Fig. 3 and is a part of the more comprehensive biological network called Hetionet [16, 17]. It can be thought of as a special case of the graph shown in Fig. 2 (Users = Compounds, Items = Diseases, $X$ = *Compound-treats-Disease*, $J_1 = L_1$ = Genes, $U_1$=*Compound-binds-Gene*, $B'_1$=*Disease-associates-Gene*). The files used in our test can be downloaded at *www.cs.uni.edu/~poleksic/hetionet_files.tar.gz*.

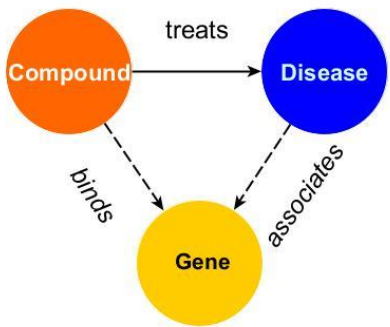

**Fig. 3.** Drug-gene-disease network.

Just like the drug-target interaction benchmark, our drug-disease benchmark uses three rounds of 5-fold cross validation to assess the prediction accuracy of three methods: EASE, MF, and DUET. The EASE method is equivalent to the previously described EASEdt algorithm (section 3.1). For improved accuracy, the MF algorithm tested here is given side information in terms of compound-compound similarity scores and disease-disease similarity scores. The similarity of two compounds is defined as the Jaccard similarity [11] of their (binary) gene profiles, where the gene profile of a compound is simply a binary vector in which a 1 represents a known gene-target interaction. The same approach is used to define and quantify pairwise similarities of diseases.

As seen in Table 2, the neighborhood-regularized matrix factorization compares favorably to the bare-bone EASE algorithm. This is somewhat expected since the latter method can only access sparse data on the compound-disease associations. However, the DUET algorithm has a significant advantage over both methods across all four different classification metrics employed in our benchmark (t-test p-value less than 0.01).

**Table 2.** Drug repurposing benchmark

|      | AUPR        | NDCG100     | PREC50      | PREC100     |
|------|-------------|-------------|-------------|-------------|
| EASE | 0.287±0.021 | 0.530±0.022 | 0.627±0.020 | 0.451±0.007 |
| MF   | 0.371±0.008 | 0.567±0.013 | 0.652±0.024 | 0.503±0.009 |
| **DUET** | **0.400±0.003** | **0.609±0.004** | **0.696±0.007** | **0.544±0.003** |

## 3.3 Speed advantage

DUET has the speed comparable to EASE which is significantly faster than algorithms based on matrix factorization. Fig. 4 shows the speed comparison between the MATLAB implementations of MF and DUET in completing a single DrugBank interaction matrix (with all hyperparameters held fixed). The speed analysis was carried out on a Ubuntu 22.04 system with 11th Gen Intel(R) Core(TM) i9-11950H @ 2.60GHz CPU and 32GB of RAM.

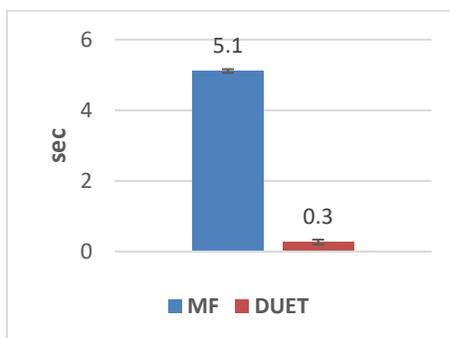

**Fig 4.** Speed comparison: DUET vs. matrix factorization.

The speed advantage of the DUET method is further pronounced in cases where the methods' hyperparameters need to be fine-tuned for optimal performance. This is because MF employs at least eight of those parameters including latent dimension, number of iterations, learning rate, etc. In contrast, DUET has only three parameters, namely the regularization parameter $\lambda_1 = \lambda_2$ and the parameters $\beta_1$ and $\gamma_1$ that quantify the role of side information in the learning process (section 2.3).

# 4 Conclusion

We present a generalized linear model for sparse matrix completion. Specifically, we show how the recently proposed shallow autoencoder can utilize multiple sources of side information to increase its ranking accuracy. Our DUET algorithm compares favorably in speed and accuracy to the widely used state-of-the art methods for predicting drug-target interactions and drug-disease associations. Further improvements can be obtained by utilizing more diverse sources of side information. In predicting drug-disease associations, for instance, those sources can include adverse reactions of drugs, drug pharmacologic classes, disease locations, and disease symptoms, to name a few.